\documentclass{article}
\usepackage{spconf,amsmath,graphicx}

\usepackage[dvipsnames]{xcolor}
\definecolor{mypink1}{rgb}{0.658, 0.088, 0.178}

\usepackage{todonotes}
\usepackage{url}
\usepackage{epsfig}
\usepackage{amssymb}
\usepackage{comment}
\usepackage{multirow}
\usepackage[np, autolanguage]{numprint}
\usepackage{enumitem}
\usepackage{hyperref}
\usepackage[utf8]{inputenc}
\usepackage{soul}
\usepackage{booktabs}

\newcommand{\om}[1]{[{\bf \color{green} Om: #1}]}

\title{Humans disagree with the IoU for measuring object detector localization error }
\name{Ombretta Strafforello$^{\star,1,2}$\thanks{$^{\star}$Authors with equal contribution.}, Vanathi Rajasekart$^{\star,1}$, Osman S. Kayhan$^{\star,1}$, Oana Inel$^{\star,\ast,3}$\thanks{$^{\ast}$Work performed while at Delft University of Technology.}, Jan van Gemert$^{1}$
}
\address{$^{1}$Delft University of Technology, Delft, the Netherlands, $^{2}$TNO, The Hague, the Netherlands, \\$^{3}$University of Zurich, Zurich, Switzerland\\{\{O.Strafforello, O.S.Kayhan, J.C.vanGemert\}}@tudelft.nl,\\ V.S.Rajasekar@student.tudelft.nl, inel@ifi.uzh.ch}
\begin{document}

\maketitle

\begin{abstract}
The localization quality of automatic object detectors is typically evaluated by the Intersection over Union (IoU) score. In this work, we show that humans
have a different view on localization quality. To evaluate this, we conduct a survey with
more than 70 participants. Results show that for localization errors with the exact same
IoU score, humans might not consider that these errors are equal, and express a preference.
Our work is the first to evaluate IoU with humans and makes it clear that relying
on IoU scores alone to evaluate localization errors might not be sufficient.
\end{abstract}


\section{Introduction}
\label{sec:introduction}

\begin{figure}[!ht]
	\centering
	\begin{tabular}{c@{}c}
	\includegraphics[width= 0.95\linewidth]
	{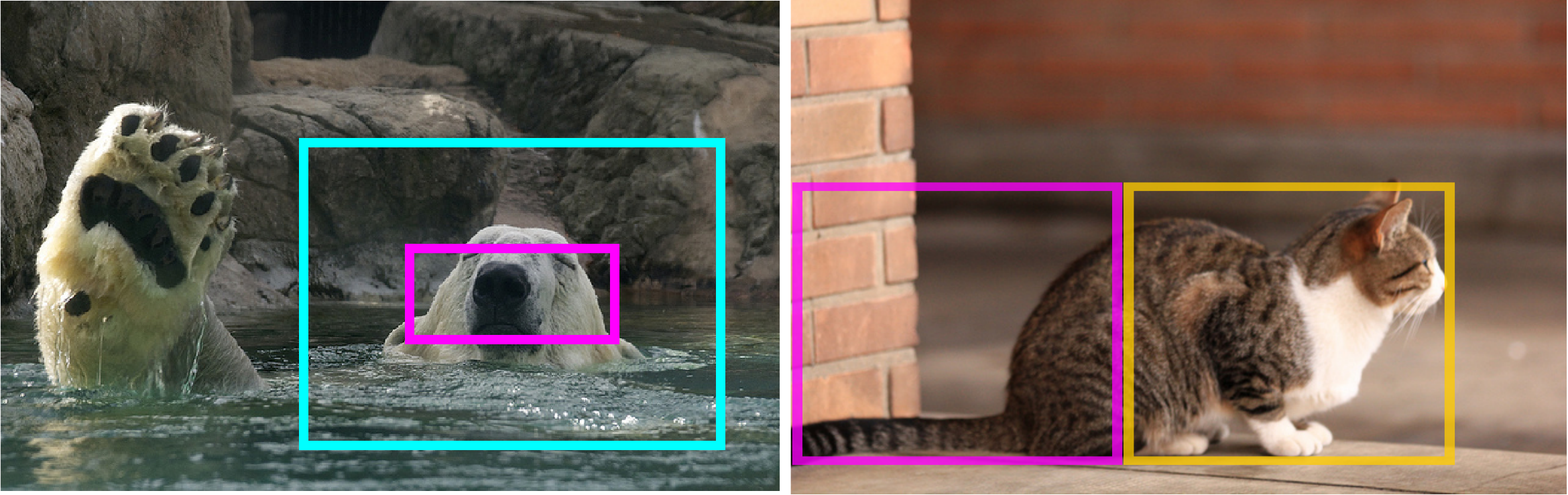}
	%{images/Acc_Recall_of_onlyabsent_parts_all29.pdf}
	\end{tabular}
	\caption{Left: Two localizations  where the magenta box ($0.5$ IoU) is accepted, and the cyan box ($0.3$ IoU) is rejected by object detectors. Right: Two equally accepted localizations ($0.5$ IoU) by object detectors. Which boxes do you accept? %Here, we perform user studies with around 70 participants to evaluate how humans perceive such cases. 
	}
	\label{fig:fig1}
\end{figure}

The main difference between image classification and object detection is that an object detector also has to predict the object's location, typically indicated by a bounding box around the object. Object location can be used as a first step for a downstream task, e.g., instance segmentation~\cite{he2017maskRCNN}, or human pose estimation~\cite{cao2019openpose}. Alternatively, in this paper, we focus on the setting where an object detection is presented to humans as an end result, where examples include visual inspection~\cite{kayhan2021hallucination}, or focusing attention in medical images~\cite{li2019clu}. We do not evaluate the object detector itself~\cite{hoiem2012diagnosing}. Instead, we evaluate if the predicted object location by object detectors aligns with what humans consider a detected object.

\textbf{Evaluating object detectors.}
Object detectors
%such as two-stage models \cite{girshick2015region, girshick2015fast, fasterrcnn}, single-stage approaches \cite{yolo, ssd, lin2017focal, redmon2018yolov3}, pointwise/anchorless methods \cite{law2018cornernet, duan2019centernet, zhou2019bottom} and transformers-based detectors \cite{Carion_2020, beal2020transformerbased, zhu2020deformable} 
are commonly evaluated~\cite{hoiem2012diagnosing,pascal-voc-2012,OpenImages,Salton86IAP, osk_context} with mean average precision (mAP): the mean of the per-class average precision scores. Average precision is the area under the precision-recall curve, created by ranking all detections by confidence and then checking if they are correct according to the ground truth. %The correctness of a detection depends on the classification: if the assigned class label is wrong, the detection is wrong. 
The detection is correct if (1) the assigned class label is correct and (2) the detection location has sufficient overlap with the ground truth. %A second criteria for correctness is that the location of the detection has sufficient overlap with the ground truth. 
The Intersection over Union (IoU) score is used to determine the overlap.  
The location of a detection is correct if the IoU score is higher than a threshold, typically 0.5 or higher~\cite{coco,pascal-voc-2012}.
% , where \cite{dollar2009pedestrian} states that: "The threshold of $50\%$ is arbitrary but reasonable". 
In this paper, to the best of our knowledge, we are the first to investigate how well the IoU measure aligns with human localization quality judgments. %We are not aware of other papers investigating this problem. %how well the IoU aligns with human evaluations of object localization quality. 

\begin{comment}
\textbf{Optimizing object detectors.} %In addition to evaluating object detectors at test-time, t
The IoU also plays a role in detector optimization during training, where the IoU is often optimized as an additional loss term~\cite{fasterrcnn}. In addition to the loss, the IoU might determine if a region is used for training as foreground or background, where for example, Faster RCNN~\cite{fasterrcnn} thresholds foreground at 0.7 and RetinaNet \cite{lin2017focal} at 0.5. IoU may also determine non-maximum suppression of spurious detections~\cite{Redmon_2017_CVPR, lin2017focal,shrivastava2016training}. Other examples explicitly use IoU in the optimization~\cite{jiang2018acquisition, wu2020iou}. In our paper, we link the IoU to human judgments, which can align the optimization towards human preferred detections.
\end{comment}

%%%%%%%%%%%%%%%%%%%%%%%%%%%%%%%%%%%%%%%%%
%%%%%%%%%%%%%%%%%%%%%%%%%%%%%%%%%%%%%%%%%
\textbf{Human annotation for object detection.}
\label{sec:rw_human_annotations}
Extensive crowdsourcing studies are performed to draw bounding boxes around objects in images \cite{zhu2012we,song2015robot}
%and videos \cite{vondrick2013efficiently} 
or the precise shape of the object \cite{yuen2009labelme,russell2008labelme}. 
Experiments in which crowd workers validate object detections showed that annotators tend to be lenient when validating bounding boxes, \emph{i.e.}, bounding boxes with IoU $<$ 0.5 are still accepted \cite{papadopoulos2016we}. 
Furthermore, analyses performed in \cite{russakovsky2015best} suggest that to efficiently and accurately localize all objects in an image, several crowdsourcing tasks are needed, such as verifying box correctness, verifying object presence, or naming the object. 
In this paper, we extend the work in \cite{su2012crowdsourcing,pang2019libra,russakovsky2015best} with four user studies investigating which bounding boxes humans accept %(\emph{i.e.}, validation) 
and prefer. % (\emph{e.g.}, comparison). 
%We experiment with various object sizes (small, large, medium), IoU values (0.3, 0.5, 0.7, 0.9), and bounding box scales (small and large) and positions (top/front and bottom/back). 

\textbf{Contributions.}
We make the following contributions: \emph{(1)} We design four user studies to explore what kind of detections humans prefer and accept as good detections.%and what boxes are accepted as good detections. 
\footnote{Data and analysis is available at \url{https://github.com/ombretta/humans_vs_IoU}.} %Upon the acceptance of the paper, we will publish all data collected and the analyses performed. 
\emph{(2)} We investigate the relationship between a too small bounding box and a too large bounding box, where they both have the same IoU score. \emph{(3)} We analyze the impact of object symmetry and bounding box position in human preference and acceptance of detectors' output. \emph{(4)} We experiment with various object sizes (small, medium, large) and recommend future studies. %preliminary results regarding the influence of participants' computer vision knowledge in the perception of bounding boxes' suitability.

Our results show that humans disagree with IoU for measuring localization errors.

\section{Experimental Approach}
\label{sec:experiment}

% - What is the motivation to do the survey?\\
% - How do we design it? \\
%     i. What do we investigate and why?\\
%     ii. Show and explain each of them\\
% - Research questions (1,2,3,4) \\
%     i. Question \\
%     ii. Setup with examples \\
%     iii. Results \\

%%%%%%%%%%%%%%%%%%%%%%%%%%%%%%
We perform four controlled experiments to evaluate the relation between IoU and human localization quality judgments and study which object detections are accepted or preferred by humans. We do not train or test any object detection models since they are highly influenced by many design choices, e.g., model parameters, dataset. Thus, our boxes are generated according to the ground truth.
%We design four user studies to evaluate object detection localization decisions by humans. The main focus of the studies is to understand how object localization is received by humans and study which object detections are accepted or preferred by humans, and 
We relate our findings to machine-evaluated detections. 
For machine-evaluated detections, we use the common IoU, measuring the localization performance of the predicted box $B_p$ with the ground truth box $B_{gt}$, as  $\text{IoU}= \frac{B_p \cap B_{gt}}{B_p \cup B_{gt}}$.

%We do not train or test any object detection models since they are highly influenced by many design choices, e.g., model parameters, dataset. We aim to discard all these factors and only focus on how humans perceive such predictions. Therefore, our boxes are generated according to the ground truth.

We address two important features of object localization:
    \textit{(i) Box Size} and
    \textit{(ii) Box Position}, which are affected by the IoU score, in four online user studies (two studies per feature).\footnote{Ethical approval was not required - we do not collect personal identifiers.}
%Both the scale and the position of the predicted box affect the IoU score. 
We also experiment with various object sizes (small - S, medium - M, large - L)\footnote{We adopt the definition of object size provided with the MS COCO dataset (\url{https://cocodataset.org/\#detection-eval}).} and IoU values (0.3, 0.5, 0.7, 0.9) to study differences and similarities between humans and detection algorithms. %We ran the studies on the Qualtrics \cite{qualtrics} tool: \emph{a)} two user studies in which we experiment with the \emph{box size} and \emph{b)} two user studies in which we experiment with the \emph{box position}. The user studies have been distributed among research group members and authors' peers.

\textbf{Procedure and participants.} All studies follow the same procedure. 
% First, the study participants are asked if they have any domain knowledge, namely computer vision knowledge. Then, participants are given an example to introduce the task.
Participants are given an example to introduce the task. 
%Each research question starts by asking if a participant has domain knowledge. We had only X number of domain expert from Y number of total participants. Afterwards, for each research question, an example question is demonstrated to introduce the task to the participants. Each question incorporates 
The task consists of a masked image to indicate which object is investigated, the question that directly specifies the object name, and the possible answers. The images are chosen from the MS COCO dataset \cite{lin2014microsoft}. We ran the studies using Qualtrics\footnote{\url{https://www.qualtrics.com/}}. The user studies have been distributed among research group members and authors' peers.
% (see in Supplementary Material Fig. S1, S5, S7 and S9). %\ref{fig:survey_sample}). 
% The question and the possible answers differ per study, as shown in Table \ref{tab:setup}. We elaborate on them below. 

\begin{figure}[!ht]
	\centering
	\begin{tabular}{c@{}c}
	\includegraphics[width= 0.95\linewidth]
	{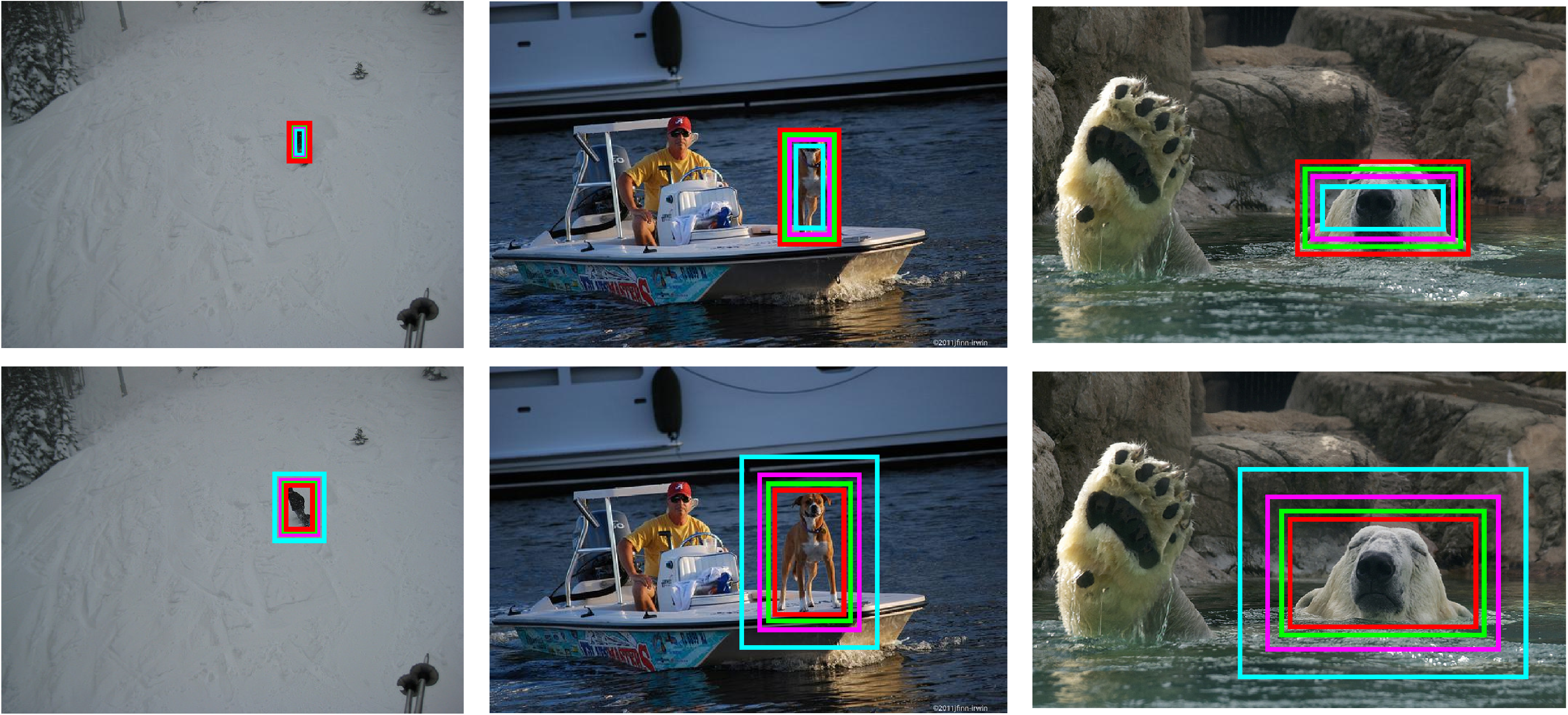}
	%{images/Acc_Recall_of_onlyabsent_parts_all29.pdf}
	\end{tabular}
	\caption{Size preference experiment. The columns indicate Small, Medium and Large object categories. %of COCO \cite{coco}. 
	The colors represent IoU scores of each box: Red ($0.9$), Green ($0.7$), Magenta ($0.5$) and Cyan ($0.3$). Top row: small bounding boxes; Bottom row: large bounding boxes. 
% 	To note that, the small and large boxes of same color have the same IoU scores, \emph{e.g.}, all the red boxes have $0.9$ IoU score. 
	The small and large boxes of same color have the same IoU scores.}
	\label{fig:scale_pref}
\end{figure}

\emph{Box Size.} As illustrated in Fig. \ref{fig:scale_pref}, we use two different box sizes, \textit{small} and \textit{large}, with the same IoU score. 
The box aspect ratio and position is taken from the ground truth box. In the \textit{Size Preference} study,
% (Fig. S1)
we investigate the box size, and ask participants which box size they prefer for a detection. They can choose one option among: large box, small box or ``the size of the box does not matter''. In the \emph{Size Acceptance} study, we show either a small or a large box and ask participants if they accept or reject it as an object detection. For both studies we evaluate IoU values (0.3, 0.5, 0.7, 0.9) and include all object sizes (S, M, L). In the \emph{Size Preference} study, we annotate 72 images, with six images per each combination between object size and IoU value. In the \emph{Size Acceptance} study, we annotate 96 images (eight per combination). % between object size and IoU value.

% , see the first row in Table \ref{tab:setup}. 
%In this study, the images were equally distributed across object sizes (small, medium, large) and IoU values $\in \{0.3, 0.5, 0.7, 0.9\}$.
%We used a total of \np{72} images, with \np{6} images per each combination between object sizes and IoU values. In the \textit{Scale Acceptance} study,
% (Fig. S5), 
%we ask participants if they accept or reject a given box as an object detection.
% (see the second row in Table \ref{tab:setup}). 
%In total, \np{96} images have been evaluated in this study.
%The study includes 8 images for each survey category of object size and IoU values.
% . The images were equally distributed across object sizes (small, medium, large) and IoU values (0.3, 0.5, 0.7, 0.9), 

%The questions about scale in the survey comprise two types of directions. In the first type, it is asked to a user if the user accepts or rejects the given object localization, see the first raw in Tab. \ref{tab:setup}.
%The other type contains the questions to obtain the user's preferences and asks whether user chooses large box, small box or the size of the box does not matter, see the second raw in Tab. \ref{tab:setup}.
%The images for both tasks are chosen \textcolor{red}{(if the number is equally chosen)} from small, medium and large object size groups and the observations are determined with IoU thresholds $0.3$, $0.5$, $0.7$ and $0.9$ (Tab. \ref{tab:setup}).

 \begin{figure}[!ht]
	\centering
	\begin{tabular}{c@{}c}
	\includegraphics[width= 0.95\linewidth]
	{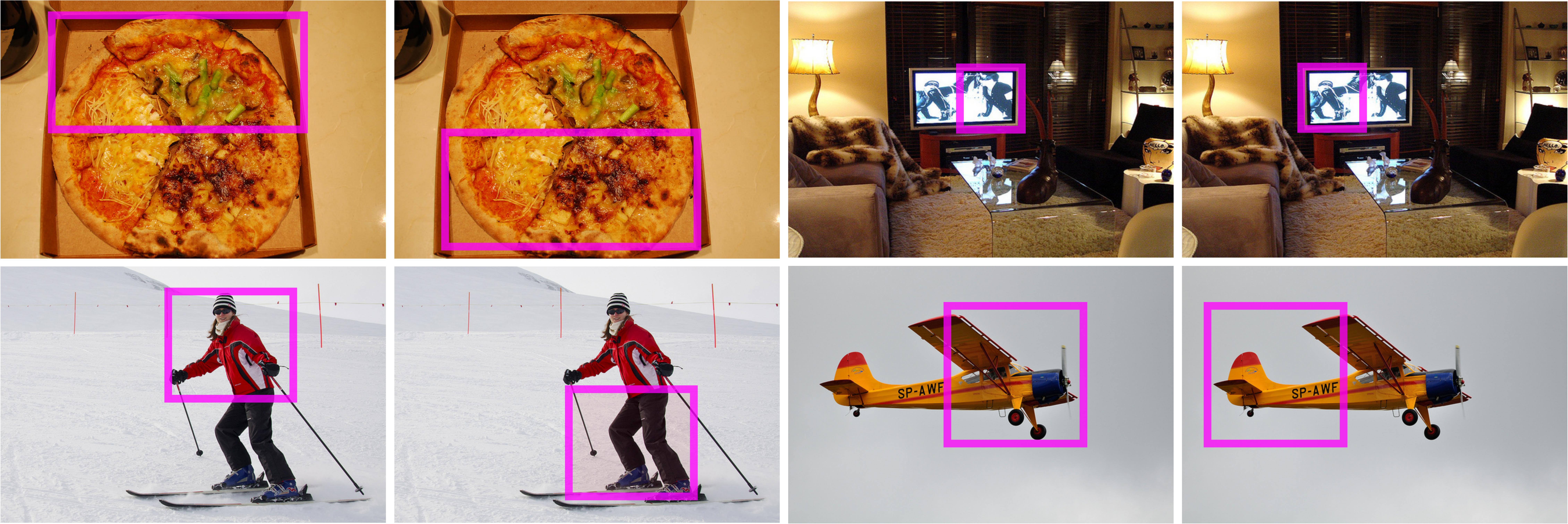}
	%{images/Acc_Recall_of_onlyabsent_parts_all29.pdf}
	\end{tabular}
	\caption{Position preference experiment. The experiments show the bounding box locations for IoU score $0.5$ by shifting them horizontally or vertically. Top row: symmetrical objects; Bottom row: asymmetrical objects. %Object detection methods accept all these boxes similarly as true positive regardless of their orientation. 
	}
	\label{fig:position_pref}
\end{figure}

\emph{Box Position.} As illustrated in Figure \ref{fig:position_pref}, we applied two positional shifts to the ground truth box, for symmetrical and asymmetrical objects, using a fixed IoU value of 0.5.
% , as the value is a threshold of correct localization for many object detectors \cite{girshick2015region, ren2015faster, yolo, redmon2018yolov3, ssd, fu2019retinamask} and datasets \cite{pascal-voc-2012, coco}. 
Unlike the size experiment, the predicted box size is fixed and only the position of the box changes to evaluate the effect of the position. Depending on the orientation of the object, the predicted box is shifted horizontally (back, front) or vertically (top, bottom). Since symmetrical objects do not have front and back sides, we consider front as the right side and back as the left side of the object.
Similarly to the size surveys, in the \textit{Position Preference} study,
% (Fig. S7), 
we ask participants if they prefer a particular part or side of the object for detection.
% (see the third row in Table \ref{tab:setup}). 
The \textit{Position Acceptance} study
% (Fig. S9) 
investigates if users would accept the bounding box as a correct detection.
% (see the fourth row in Table \ref{tab:setup}). 
In both position surveys, we use \np{20} images, which are equally distributed across object types (symmetrical, asymmetrical) and box positions (front/top, back/bottom), with 5 images per category.
%In total, we also use \np{20} images, namely 5 images per survey category.  

% Similarly to the scale surveys, the first box position survey investigates users' acceptance regarding the position of the bounding box (see the third row in Table \ref{tab:setup}). In total, we used \np{20} images, which were equally distributed across object types (symmetric, asymmetric) and box positions (front/top, back/bottom), namely 5 images per category. In the second box position survey, we ask participants if they prefer a particular part or side of the object for detection (see the fourth row in Table \ref{tab:setup}). In this survey, we again used \np{20} images, with 5 images per category. 

%include users' acceptance and preference questions. accept/reject type questions and user's preferences questions. %For position experiments, the IoU value is fixed on 0.5 IoU score as the value is a threshold of correct localization for many object detectors \cite{girshick2015region, ren2015faster, yolo, redmon2018yolov3, ssd, fu2019retinamask} and datasets \cite{pascal-voc-2012, coco}.
%The aim of the position experiments is to check whether humans accept (Tab. \ref{tab:setup}, 3rd raw) or prefer (Tab. \ref{tab:setup}, 4th raw) a particular part or side of the object for detection. Actually, the position of the predicted box does not matter for object detectors when the IoU score is the same, namely a detector will always accept the localization of the object if the IoU is equal or greater than 0.5. 

\section{Results}
\label{sec:results}

\begin{figure}[!ht]
\centering
\begin{tabular}{c}
\includegraphics[width=6.5cm]{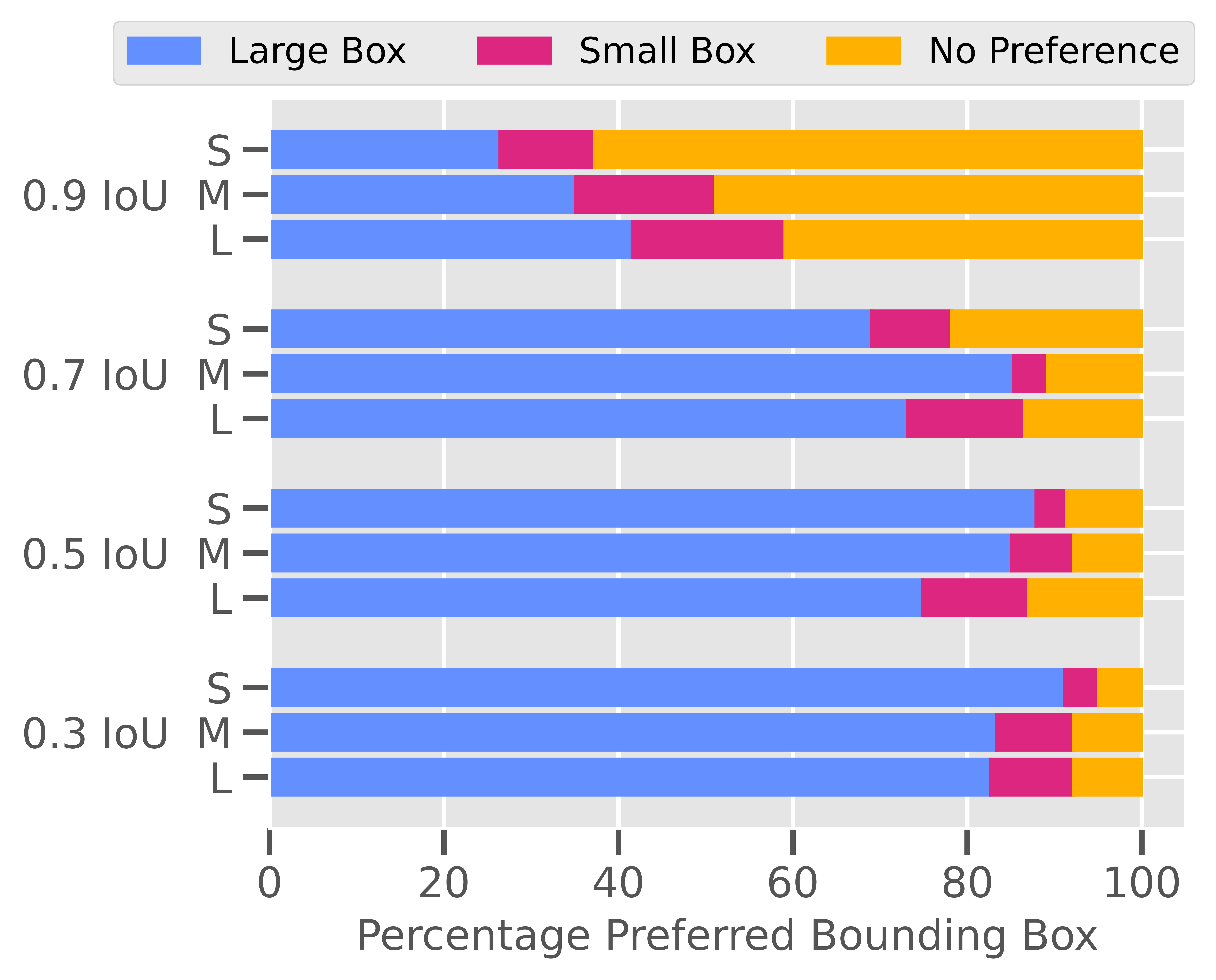}\\ 
(a) Size Preference Study\\
\includegraphics[width=6.5cm]{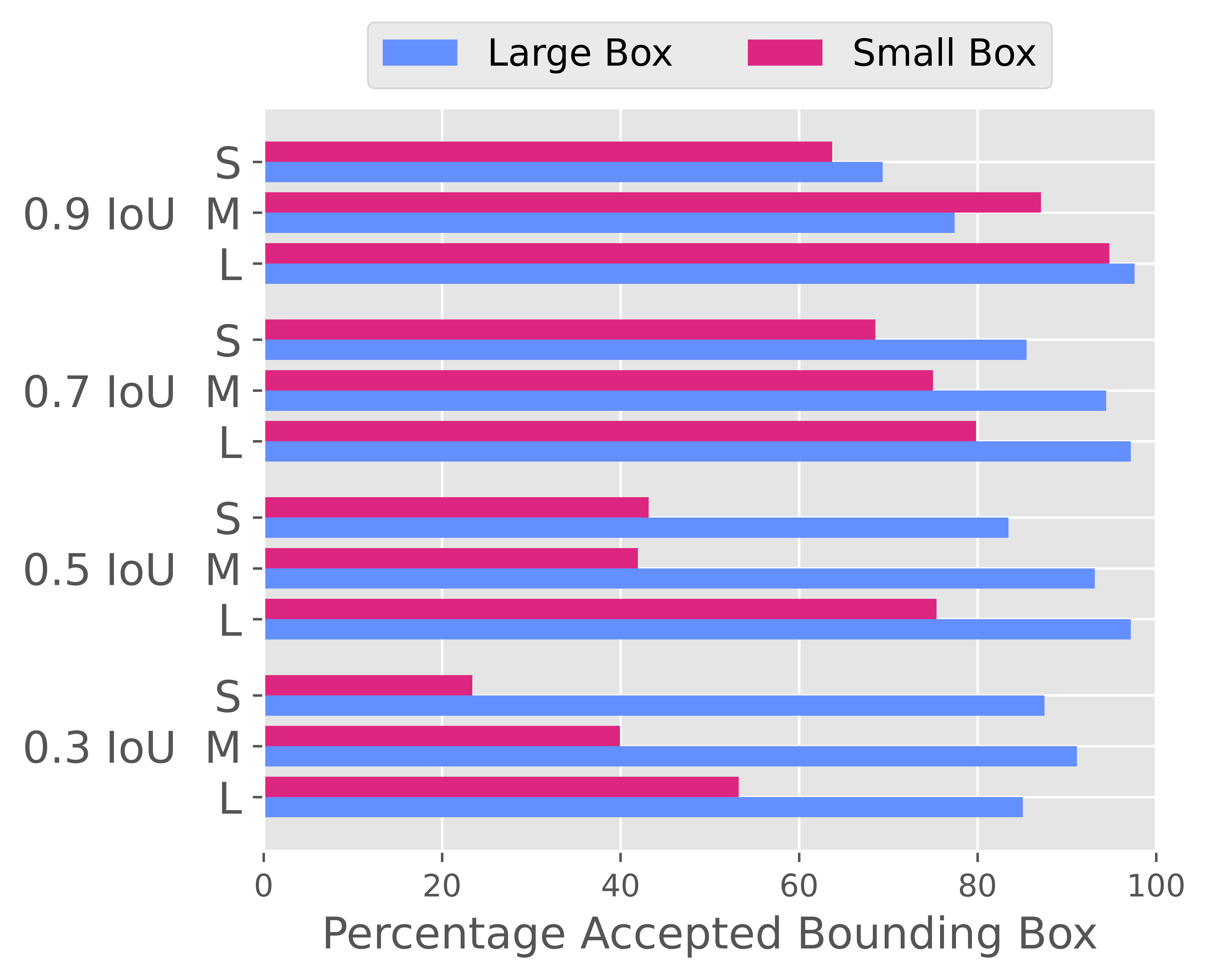}\\
(b) Size Acceptance Study
\end{tabular}
\caption{Results from studies \emph{Size Preference} and \emph{Size Acceptance}. a) Percentage of preferred bounding box size (small, large, no preference) for each IoU (0.3, 0.5, 0.7, 0.9) and object size (S, M, L). b) Percentage of accepted bounding box size (small, large) for each IoU and object size. The large boxes are mostly preferred and accepted by humans.}
\label{fig:survey1and2}
\end{figure}

\textbf{Analytical method.}
To study the human preference and acceptance of bounding box sizes and positions, we apply several statistical tests. We apply the Chi-square test \cite{mchugh2013chi} to find out if there are any associations between variables such as object size and preferred box size or IoU value and preferred box size. 
% The Chi-square test determines whether there is a relationship between studied  variables. 
To understand whether differences in preference proportions (\emph{e.g.}, small boxes, large boxes, no preference), or acceptance proportions (\emph{e.g.}, front box, back box) are statistically significant, we apply the Z-test \cite{schumacker2017z} and the Cochran's Q test \cite{cochran1950comparison}. While the Z-test can only be applied to compare two proportions, the Cochran's Q test can be applied on any number of proportions. 
%The Cochran's Q test, however, does not indicate which proportion is statistically significantly different from the rest. Thus, 
In case of statistically significant differences, we apply a posthoc Dunn test with Bonferroni correction \cite{weisstein2004bonferroni} to see which proportions are different. %The posthoc Dunn test is used to perform pairwise comparisons between each two proportions, while the Bonferroni correction is needed due to multiple pairwise comparisons. 
Since for each study we perform multiple comparisons and statistical tests, we use a lower significance threshold than 0.05 (by applying a Bonferroni correction), \emph{i.e.}, $\alpha$ = $\frac{0.05}{\#tests}$. %do not handle the typical significance threshold of 0.05. Instead, w
%We apply a Bonferroni correction and we use a lower significance threshold, depending on the number of tests we apply on each study, \emph{i.e.}, $\alpha$ = $\frac{0.05}{\#tests}$. % = corrected\_alpha$.
% (see Supplementary Material for the corrected significance values).

\textbf{Size Preference.} Figure \hyperref[fig:survey1and2]{\ref{fig:survey1and2}(a)} shows, per IoU and object size, the percentage of preferred bounding box sizes. For 0.9 IoU value, people have no size preference --- for each object size, the option \emph{no preference} is either the most chosen, or similarly chosen as \emph{large boxes}. 
% For each IoU value and each object size, we applied the Cochran's Q test \cite{cochran1950comparison} to see whether there are significant difference in participants' preferences regarding box scales. 
For IoU values of 0.9, posthoc Dunn tests with Bonferroni correction show that \emph{no preference} is statistically preferred for small and medium objects, but not for large objects. %, but it is statistically significantly preferred for small and medium objects.
The prevalence of \emph{no preference} is sensible: for $\text{IoU}>0.9$, the difference in appearance between small and large boxes is subtle to the human eye.

\begin{figure}[!ht]
\centering
\begin{tabular}{c}
\includegraphics[width=6.5cm]{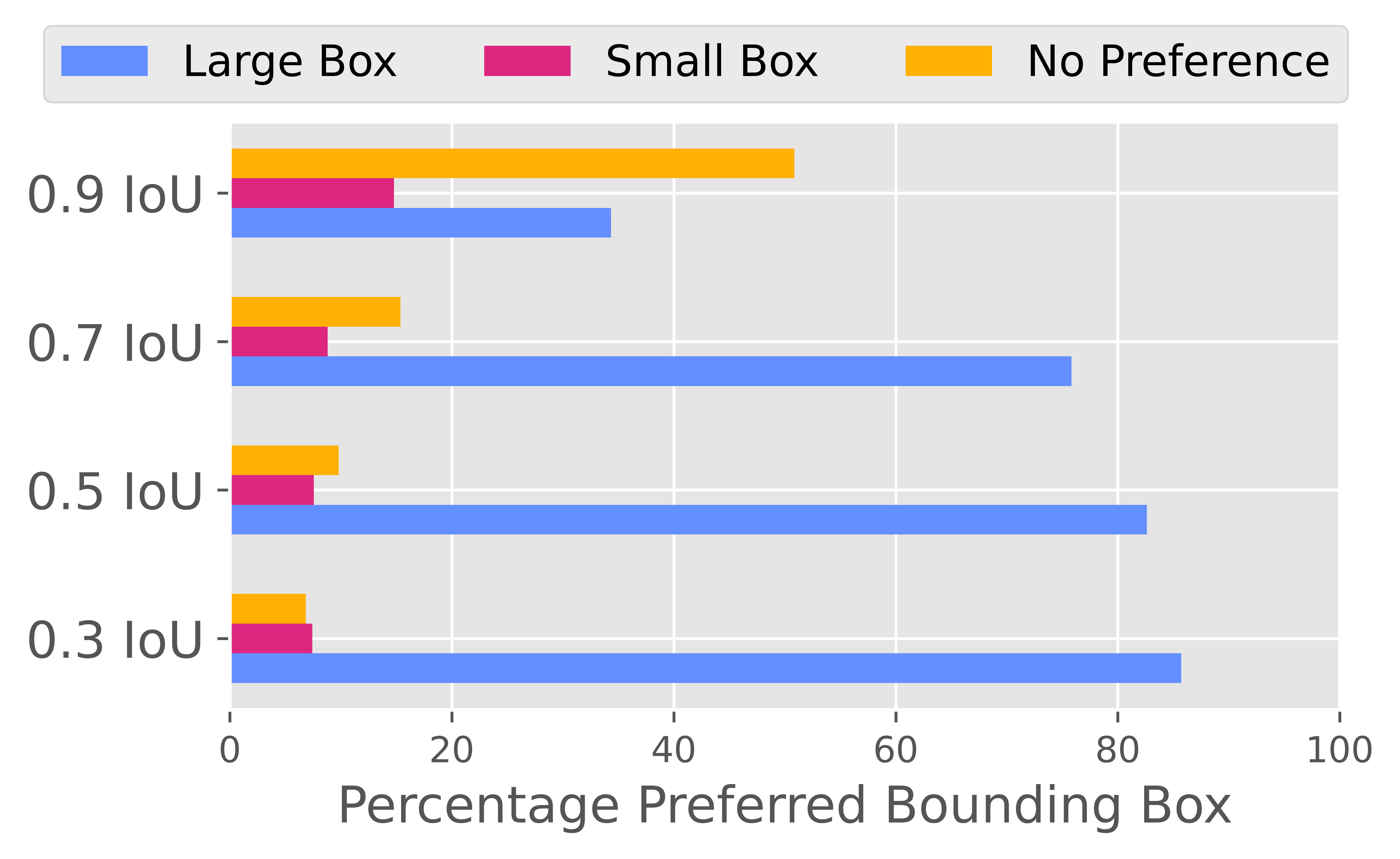}\\
(a) IoU Value vs. Bounding Box Size\\
\includegraphics[width=6.5cm]{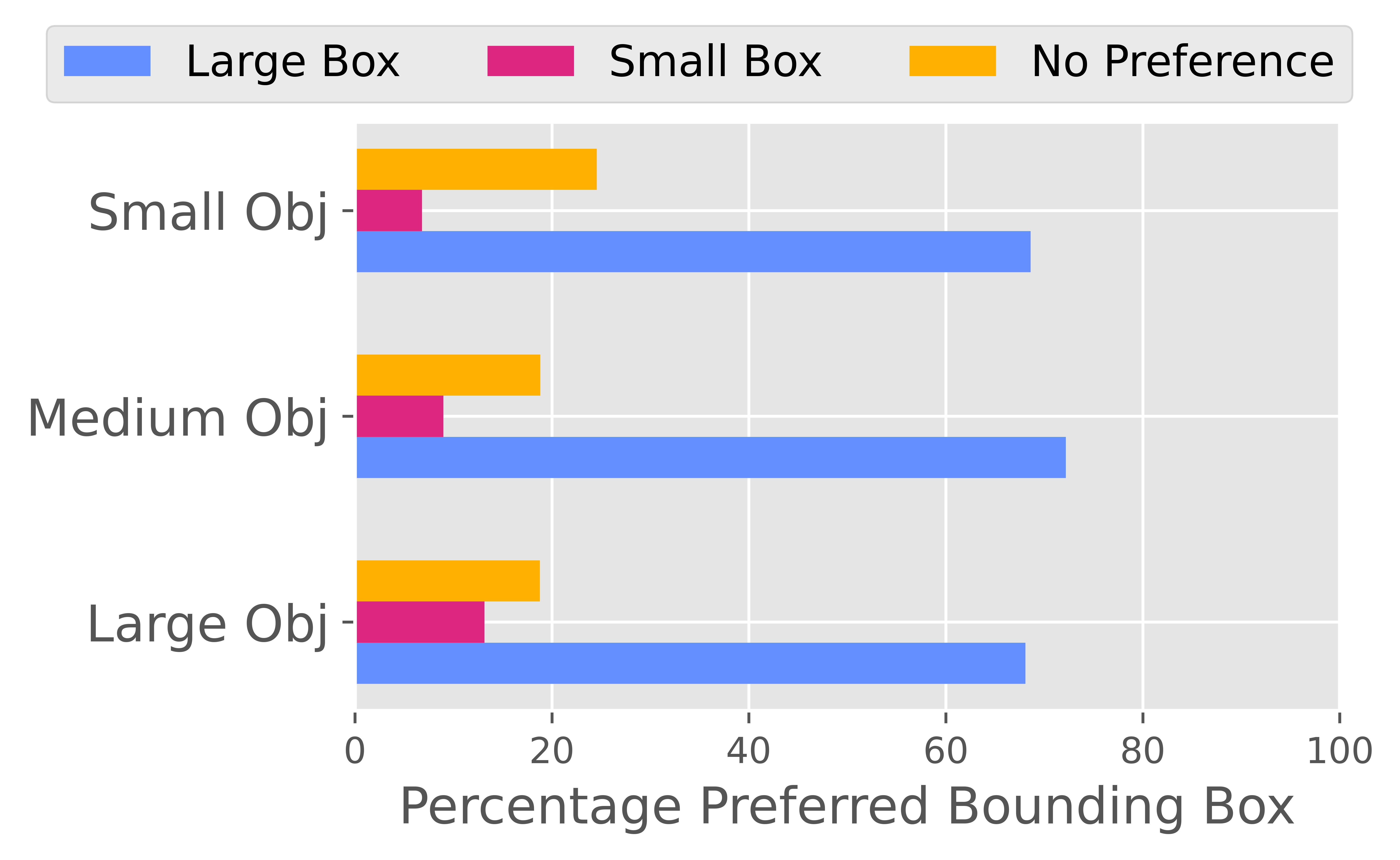}\\
(b) Object Size vs. Bounding Box Size
\end{tabular}
\caption{Results from \emph{Size Preference} study. a) Percentage of preferred bounding box size (small, large, no preference) for each IoU value (0.3, 0.5, 0.7, 0.9). b) Percentage of preferred bounding box size for each object size (S, M, L).}
\label{fig:survey1_associations}
\end{figure}

For all other evaluated IoU values, 0.7, 0.5, 0.3, and for all three evaluated object sizes, the Cochran's Q test shows that there are statistically significant differences in the preference of boxes. Posthoc Dunn tests with Bonferroni correction indicate that \emph{large boxes} are statistically significantly more preferred by humans.
%\textbf{Is there an association between the bounding box scale and the object size?}
% In the \emph{Scale Preference} user study, for all object sizes, we observe that \emph{large} bounding boxes are always preferred. 
Small bounding boxes are always the least preferred while large bounding boxes are always the most preferred, irrespective of object size. 
We observe a gradual preference increase of \emph{small} bounding boxes as the IoU value increases, and a comparatively higher increase in having \emph{no preference} (see Figure \hyperref[fig:survey1_associations]{\ref{fig:survey1_associations}(a)}). Using a Chi-square test, we found an association between the IoU value and the preferred bounding box size ($\chi^2$(2)=1227.84, $p<0.006$). We also notice a gradual decrease in the preference of small bounding boxes with the decrease of the object size. These results are shown in Figure \hyperref[fig:survey1_associations]{\ref{fig:survey1_associations}(b)}. Using a Chi-square test, we found a statistically significant association between the object size and the size of the preferred bounding box ($\chi^2$(2)=62.05, $p<0.006$).

%Small bounding boxes are always the least preferred, irrespective of the object size. In addition, we also notice a gradual decrease in the preference of small bounding boxes with the decrease of the object size. These results are shown in Figure \hyperref[fig:survey1_associations]{\ref{fig:survey1_associations}(b)}. Using a Chi-square test, we found a statistically significant association between the size of the object and the scale of the preferred bounding box ($\chi^2$(2)=62.05, $p<0.006$).
%\textbf{Is there an association between the bounding box scale and the IoU value?}

%\emph{Large} bounding boxes are the most preferred for all IoU values, except for 0.9, where the option \emph{no preference} is chosen the most. We also observe a gradual increase in preference of \emph{small} bounding boxes as the IoU value increases, and a comparatively higher increase in having \emph{no preference} (see Figure \hyperref[fig:survey1_associations]{\ref{fig:survey1_associations}(a)}). Using a Chi-square test, we found an association between the IoU value and the preferred bounding box size ($\chi^2$(2)=1227.84, $p<0.006$).

\textbf{Size Acceptance.} In Figure \hyperref[fig:survey1and2]{\ref{fig:survey1and2}(b)}, we show the percentage of accepted \emph{small} and \emph{large} boxes, for each IoU value and image size. %, as resulted from the \emph{Accepted Box Scale} study. We make several observations. 
For each IoU value, the acceptance of \emph{small} bounding boxes decreases with the decrease of object size, the smaller the object, the less accepted the \emph{small} bounding boxes. \emph{Large} bounding boxes are always more accepted than \emph{small} bounding boxes, disregarding IoU values and object sizes. The exception are medium objects with 0.9 IoU, where \emph{small} boxes are statistically significantly more accepted (z=-2.82, $p<0.008$). For the rest of the cases, \emph{large} bounding boxes are statistically significantly more accepted than \emph{small} bounding boxes for IoU values of 0.3, 0.5 and 0.7 and all object sizes ($p<0.008$), but are not more accepted for neither small nor large objects with 0.9 IoU. %, according to the Z-test ($p>0.008$). %\cite{ztest,schumacker2017z}.
%We wonder whether the acceptance of \emph{small} and \emph{large} bounding boxes is statistically significantly different than their rejection. After applying the Z-test of proportions, 
We also found, c.f. Z-test, that \emph{(1)} \emph{large} bounding boxes are always statistically significantly accepted ($p < 0.008$) and \emph{(2)} \emph{small} bounding boxes are only statistically significantly more accepted for 0.9 and 0.7 IoU (all object sizes) and large objects with 0.5 IoU. % ($p < 0.008$).   

\begin{figure}
\centering
\begin{tabular}{c}
\includegraphics[width=7cm]{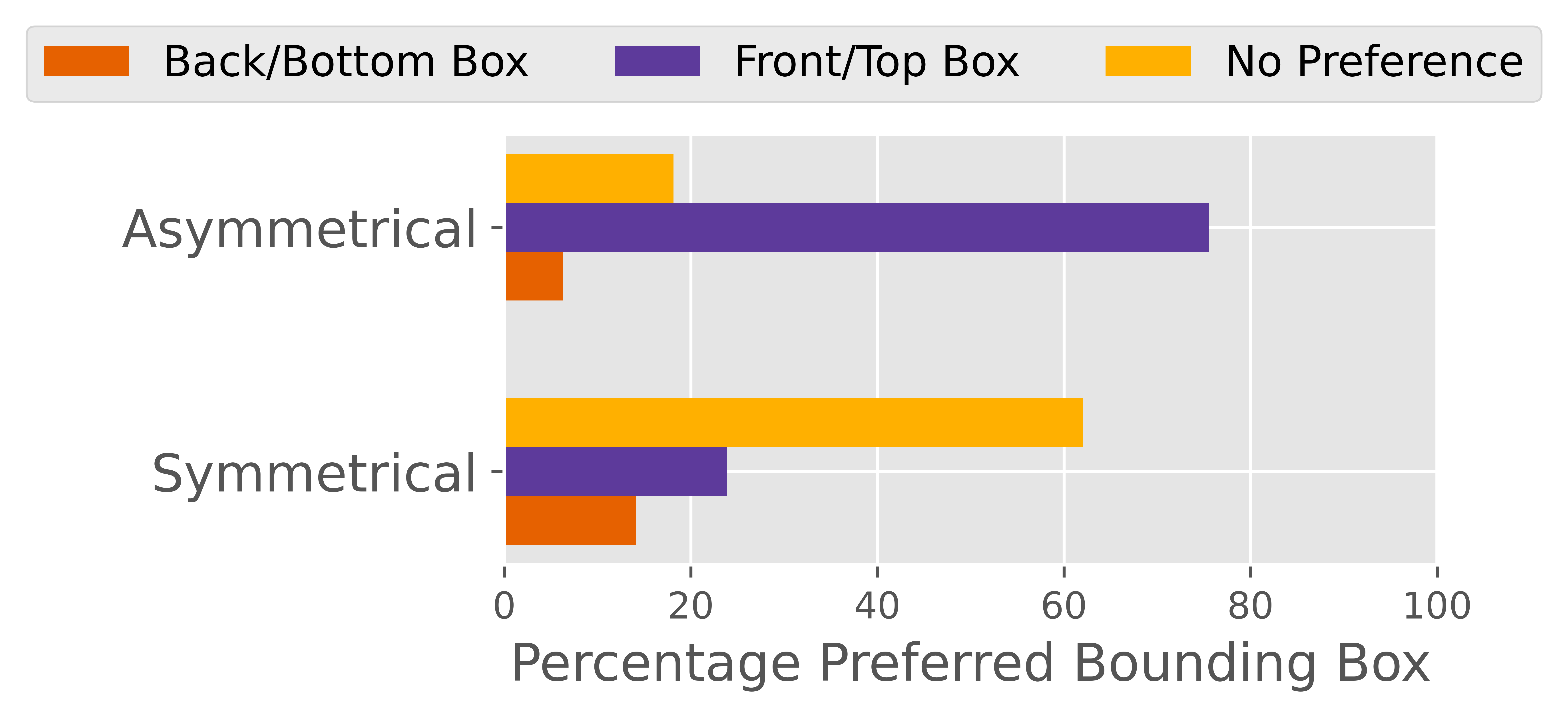}\\
(a) Preferred Box Position\\
\includegraphics[width=6.5cm]{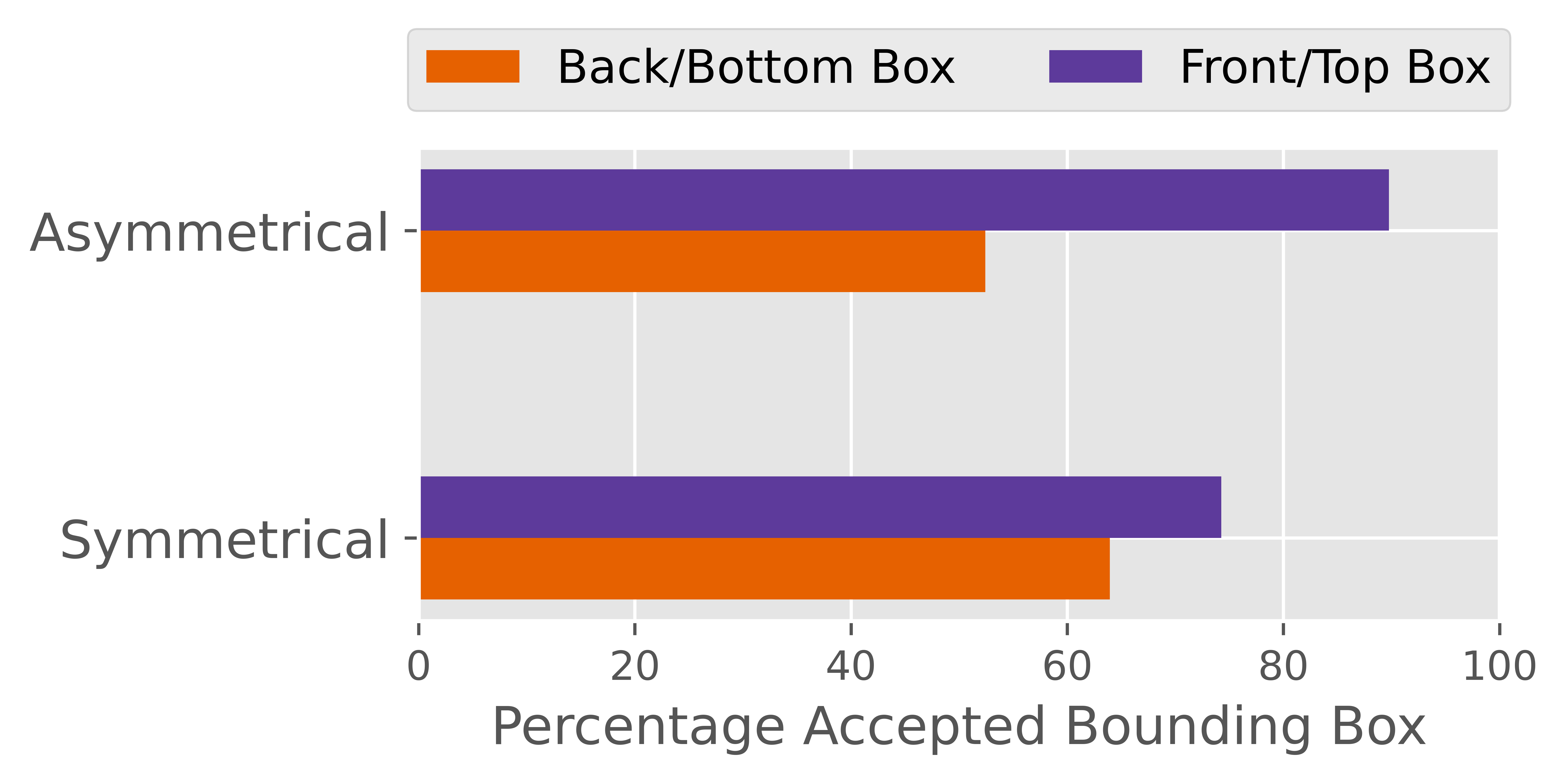}\\
(b) Accepted Box Position
\end{tabular}
\caption{Results from studies \emph{Position Preference} and \emph{Position Acceptance}. a) Percentage of preferred bounding box position (front, back, no preference) for symmetrical and asymmetrical objects. b) Percentage of accepted bounding box position (front, back) for symmetrical and asymmetrical objects.}
\label{fig:sym_asym_boxes}
\end{figure}

\textbf{Position Preference.}
Figure \hyperref[fig:sym_asym_boxes]{\ref{fig:sym_asym_boxes}(a)} presents the results of the \emph{Position Preference} user study. For symmetrical objects, participants have no preference regarding the position (\emph{front/top} or \emph{back/bottom}) of the bounding box, \emph{no preference} being chosen the most. According to the Cochran's Q test, we also find that there are statistically significant differences in proportions among the three options chosen by study participants ($\chi^2$(2)=268.76, $p<<0.017$). A pairwise posthoc Dunn test with Bonferroni correction indicates that there are statistically significant differences between the proportions in which \emph{no preference} and \emph{front} bounding boxes are preferred ($p<<0.017$), as well as between the proportions of \emph{no preference} and \emph{back} bounding boxes ($p<<0.017$). 

For asymmetrical objects, however, the most preferred bounding box is positioned at the \emph{front} of the object. %For both symmetrical and asymmetrical objects, the \emph{back/bottom} bounding box seems to be the least preferred. 
The Cochran's Q test shows that the difference in proportions among the three options is statistically significant ($\chi^2$(2) = 576.74, $p<<0.017$). Posthoc analysis using the Dunn test with Bonferroni correction shows that these differences are statistically significant between each two possible answers (\emph{front} and \emph{no preference}, \emph{front} and \emph{back}).

\textbf{Position Acceptance.}
Figure \hyperref[fig:sym_asym_boxes]{\ref{fig:sym_asym_boxes}(b)} presents the results of the \emph{Accepted Box Position} study. For both symmetrical and asymmetrical objects, the \emph{front} bounding box is accepted in higher proportions than the \emph{back} bounding box. %, above 74\%, compared to the \emph{back/bottom} bounding box. While for symmetrical objects the \emph{back/bottom} bounding box is accepted in around 64\% of the total amount of judgments, for asymmetrical objects the acceptance is only 52\%. 
%To check whether \emph{back/bottom} and \emph{front/top} bounding boxes are statistically significant accepted for symmetrical and asymmetrical objects, we apply the Z-Test of proportions. 
For symmetrical objects, we found sufficient evidence, c.f. Z-test, that the proportion of \emph{back} (z = -7.16, $p < 0.008$) and \emph{front} (z = -12.62, $p < 0.008$) bounding boxes of being accepted is higher than the proportion of not being accepted. For asymmetrical objects, however, only \emph{front} bounding boxes are statistically significant accepted (z = -20.18, $p < 0.008$).
% , while \emph{back/bottom} bounding boxes are not (z = -1.25, $p > 0.008$).
Similarly, for each object type, we analyze whether one type of bounding boxes is more accepted than the other. 
%For this, we apply again the Z-Test of proportions. 
For both symmetrical and asymmetrical objects, the \emph{front} bounding boxes are statistically significant more accepted than \emph{back} bounding boxes. % (symmetrical: z = -2.89, $p < 0.008$; asymmetrical: z = -10.49, $p < 0.008$).

\section{Discussion}
\label{sec:discussion}

%\oana{in the abstract we say something about proposing an asymmetric loss function, and I think we should mention it again here}

In this paper, we performed four user studies to understand which object detections are preferred and accepted by humans. We addressed two main features of object localization, namely the scale (large, small) and the position (front/top, back/bottom) of the bounding boxes, and we experimented with objects of various sizes (small, medium, large) and symmetries (symmetrical and asymmetrical). 

Our studies show a statistically significant relationship between the IoU value and the preferred bounding box size, as well as between the object size and the preferred bounding box size.  
% On the one hand, as the IoU score increases, small bounding boxes are less preferred while large bounding boxes are more preferred. Besides, as the size of objects decreases, the preference for small bounding boxes sees a slight increase. 
% However, the preference for large bounding boxes is consistently high across object sizes. 
%\emph{Large} bounding boxes are both the most preferred and the most accepted, while object detectors accept and prefer large and small boxes similarly if the boxes have the same IoU scores. 
%  \oana{is there something here that contradicts existing approaches?} \jvg{Well, 'normal' methods accept the both equally.}
% We also found that for symmetrical objects, the shift in the bounding box position does not affect study participants’ preference or acceptance.
% , \emph{i.e.}, they remain high for both shifts. 
% However, for asymmetrical objects, the position of the bounding box matters for study participants since they tend to choose bounding boxes that define or help them identify the object. This observation is in contrast to current state-of-the-art object localization models \cite{girshick2015region, girshick2015fast, fasterrcnn, yolo, ssd, lin2017focal, redmon2018yolov3} where all bounding box positions are considered correct, regardless of their orientation when the IoU is higher than the threshold, e.g., $0.5$.
\emph{Large} bounding boxes are both the most preferred and the most accepted, while object detectors accept and prefer large and small boxes similarly if the boxes have the same IoU scores. We also found that for asymmetrical objects, the position of the bounding box matters for study participants, since they tend to choose bounding boxes that define or help them identify the object. This observation contrasts current state-of-the-art object localization models \cite{girshick2015region, girshick2015fast, fasterrcnn, yolo, ssd, lin2017focal, redmon2018yolov3}, where all bounding box positions are considered correct, regardless of their orientation, when the IoU is higher than the threshold. %, e.g., $0.5$.

Object detection models, when intended for humans, should be developed in a user-centric manner \emph{i.e.}, they should incorporate end-users preferences and comply with end-users needs. Thus, future studies should focus more on understanding which aspects of the objects should be captured by bounding boxes. The current study can also be extended by considering multiple datasets, occluded or truncated objects or images with multiple objects, as well as bounding boxes that are not centered, or which are shifted in random positions. Nevertheless, future studies should consider improving object detectors based on human preferences. %insights gathered from humans.

%\clearpage

\bibliographystyle{IEEE}
\bibliography{refs}

\end{document}

\maketitle

\begin{keywords}

\end{keywords}

\section{Experimental approach}

Table \ref{tab:setup} provides an overview of the possible question and answers for each study.

\begin{table*}[]
\caption{Survey setup. We investigate object localization for Scale (first row) and Position (second row) and ask humans their preference and their acceptance of an object localization.}
\label{tab:setup}
\resizebox{1\textwidth}{!}{
\begin{tabular}{llllll}
\toprule
\multicolumn{1}{l}{} & Experiment & Box Setup & IoU & Question & Answer \\ \midrule
\multirow{6}{*}{Scale} & \multirow{3}{*}{Preference} & \multirow{3}{*}{\begin{tabular}[c]{@{}l@{}}Large vs.\\ Small\end{tabular}} & \multirow{3}{*}{\begin{tabular}[c]{@{}l@{}}\{0.3, 0.5, \\ 0.7, 0.9\}\end{tabular}} & \multirow{3}{*}{\begin{tabular}[c]{@{}l@{}}Which green box do you think \\ best identifies the object shown \\ in the images below?\end{tabular}} & \multirow{3}{*}{\begin{tabular}[c]{@{}l@{}}Small, Large,\\ No preference\end{tabular}} \\
 & & & & & \\  & & & & & \\ \cmidrule{2-6}
 & \multirow{3}{*}{Acceptance} & \multirow{3}{*}{\begin{tabular}[c]{@{}l@{}}Large vs.\\ Small\end{tabular}} & \multirow{3}{*}{\begin{tabular}[c]{@{}l@{}}\{0.3, 0.5, \\ 0.7, 0.9\}\end{tabular}} & \multirow{3}{*}{\begin{tabular}[c]{@{}l@{}}Do you think the green box \\ is sufficient to identify the object \\ in the image below?\end{tabular}} & \multirow{3}{*}{Yes, No} \\
 & & & & & \\  & & & & & \\ \midrule
\multirow{6}{*}{Position} & \multirow{3}{*}{Preference} & \multirow{3}{*}{\begin{tabular}[c]{@{}l@{}}Front vs. \\ Back \end{tabular}} & \multirow{3}{*}{0.5} & \multirow{3}{*}{\begin{tabular}[c]{@{}l@{}}Which green box would you prefer\\  to identify the object in the image \\ shown below?\end{tabular}}         & \multirow{3}{*}{\begin{tabular}[c]{@{}l@{}}Front, Back,\\ No preference\end{tabular}}  \\
 & & & & & \\  & & & & & \\ \cmidrule{2-6}
 & \multirow{3}{*}{Acceptance} & \multirow{3}{*}{\begin{tabular}[c]{@{}l@{}}Front vs. \\ Back\end{tabular}}    & \multirow{3}{*}{0.5} & \multirow{3}{*}{\begin{tabular}[c]{@{}l@{}}Do you accept that the green box \\ is sufficient to identify the object \\ in the image below?\end{tabular}} & \multirow{3}{*}{Yes, No} \\
 & & & & & \\  & & & & & \\ \bottomrule                                                                              
\end{tabular}
}
\end{table*}

\section{Results}

Table \ref{tab:overview_annotations} shows, for each user study, the total number of participants, with and without a computer vision background, and the total number of judgments gathered. 

\begin{table*}[!ht]
\centering
\caption{Overview of participants and their judgments we collected in the four user studies.}
\label{tab:overview_annotations}
\resizebox{1\textwidth}{!}{
\begin{tabular}{lcccccccc} \toprule
\multicolumn{1}{c}{\multirow{2}{*}{Study}} & \multicolumn{3}{c}{Participants} & & \multicolumn{3}{c}{Judgments} \\ \cmidrule{2-4} \cmidrule{6-8}
\multicolumn{1}{c}{} &  Total & \begin{tabular}[c]{@{}c@{}}w/ CV \\ Background\end{tabular} &  \begin{tabular}[c]{@{}c@{}}w/o CV \\ Background\end{tabular} & & Total & \begin{tabular}[c]{@{}c@{}}w/ CV \\ Background\end{tabular} & \begin{tabular}[c]{@{}c@{}}w/o CV \\ Background\end{tabular} \\ \midrule
Scale Preference & \np{77} & \np{48} & \np{29} & & \np{5544} & \np{3456} & \np{2088} \\ 
Scale Acceptance & \np{62} & \np{37} & \np{25} & & \np{5952} & \np{3552} & \np{2400} \\ 
Position Preference & \np{70} & \np{44} & \np{26} & & \np{1400} & \np{880} & \np{520} \\ 
Position Acceptance & \np{66} & \np{40} & \np{26} & & \np{1320} & \np{800} & \np{520} \\ \bottomrule
\end{tabular}
}
\end{table*}

\textbf{Influence of participants' background}
The complete analysis of the influence of participants' background is found in the Supplementary Material, and we summarize as follows. In the scale preference user study, we observe that participants without computer vision expertise are more lenient and more often have \emph{no preference} in terms of bounding box scale or position, compared to participants with computer vision expertise. For both types of participants, the differences in preference of bounding boxes are statistically significant. In terms of acceptance, participants without computer vision expertise tend to also accept more often \emph{small} bounding boxes, even for very low IoU values such as 0.3. While this acceptance is not statistically significant, it shows that potential users of the system are more prone to interpret object detections differently than expected. Compared to analyzing the entire dataset of judgments, when looking in particular at participants with and without computer vision expertise, we observe that, even though for medium objects with an IoU of 0.9, \emph{small} boxes are more accepted than \emph{large} boxes, this difference is not statistically significant.

The analyses performed on the judgments provided by participants with and without a computer vision background showed similar and consistent results across participant pools regarding the preference of bounding box positions. However, for symmetrical objects, we observe that the difference in acceptance proportions between \emph{front/top} and \emph{back/bottom} bounding boxes is not statistically significant, for neither group of participants.

%\clearpage

\bibliographystyle{IEEE}
\bibliography{refs}